\documentclass[sigconf,natbib=true,anonymous=false]{acmart}
\AtBeginDocument{%
  \providecommand\BibTeX{{%
    \normalfont B\kern-0.5em{\scshape i\kern-0.25em b}\kern-0.8em\TeX}}}

\setcopyright{acmcopyright}
\copyrightyear{2022}
\acmYear{2022}
\acmDOI{XXXXXXX.XXXXXXX}

\acmConference[SIGIR '22]{45th International ACM SIGIR Conference on Research and Development in Information Retrieval}{July 11--15, 2021}{Madrid, USA}
\acmBooktitle{SIGIR '22: 45th International ACM SIGIR Conference on Research and Development in Information Retrieval,
 July 11--15, 2022, Madrid, USA} 
\acmPrice{15.00}
\acmISBN{978-1-4503-XXXX-X/18/06}

\usepackage{caption}
\usepackage{graphicx}
\usepackage{capt-of}
\usepackage{enumitem}
\setlist[itemize]{leftmargin=5mm}
\setlist[enumerate]{leftmargin=5mm}

\usepackage{amssymb}
\usepackage{pifont}
\newcommand{\cmark}{\ding{51}}%
\newcommand{\xmark}{\ding{55}}%
\usepackage{times}
\usepackage{latexsym}
\usepackage{graphicx}
\usepackage{enumitem}
\usepackage{kantlipsum}
\usepackage{caption}
\usepackage[T1]{fontenc}
\usepackage[utf8]{inputenc}
\usepackage{microtype}
\usepackage{array, makecell}
\usepackage{multirow}
\newcolumntype{P}[1]{>{\centering\arraybackslash}p{#1}}
\newcolumntype{C}{>{\centering\arraybackslash\leavevmode}p{\ColWidthNormal}}

\begin{document}

\title{ArchivalQA: A Large-scale Benchmark Dataset for Open Domain Question Answering over Historical News Collections}

\author{Jiexin Wang}
\affiliation{%
  \institution{Kyoto University}
  \country{Japan}}
\email{wang.jiexin.83m@st.kyoto-u.ac.jp}

\author{Adam Jatowt}
\affiliation{%
  \institution{University of Innsbruck}
  \country{Austria}}
\email{adam.jatowt@uibk.ac.at}

\author{Masatoshi Yoshikawa}
\affiliation{%
  \institution{Kyoto University}
  \country{Japan}}
\email{yoshikawa@i.kyoto-u.ac.jp}

\renewcommand{\shortauthors}{Wang, et al.}

\begin{abstract}
In the last few years, open-domain question answering (ODQA) has advanced rapidly due to the development of deep learning techniques and the availability of large-scale QA datasets. However, the current datasets are essentially designed for synchronic document collections (e.g., Wikipedia). Temporal news collections such as long-term news archives spanning decades are rarely used in training the models despite they are quite valuable for our society. To foster the research in the field of ODQA on such historical collections, we present ArchivalQA, a large question answering dataset consisting of 532,444 question-answer pairs which is designed for temporal news QA. We divide our dataset into four subparts based on the question difficulty levels and the containment of temporal expressions, which we believe are useful for training and testing ODQA systems characterized by different strengths and abilities. The novel QA dataset-constructing framework that we introduce can be also applied to generate high-quality, non-ambiguous questions over other types of temporal document collections\footnote{The core part of the ArchivalQA dataset and its four sub-datasets are available at \url{https://tinyurl.com/ArchivalQA}, and other resources will be publicly available after the publication, including filtered ambiguous questions, other related training datasets and the code of the entire framework.}.
\end{abstract}

\ccsdesc[500]{Information systems~Content analysis and feature selection}
\ccsdesc[500]{Information systems~Digital libraries and archives}

\keywords{datasets, question answering, question generation}
\maketitle

\section{Introduction}
With the application of digital preservation techniques, more and more past news articles are being digitized and made accessible online. This results in the availability of large news archives spanning multiple decades. They offer immense value to our society, contributing to our understanding of different time periods in the history and helping us to learn about the details of the past \cite{Korkeamaki:2019:IDD:3295750.3298931}. However, due to their large sizes and complexities, it is difficult for users to effectively utilize such temporal news collections. A reasonable solution is to use open-domain question answering (ODQA), which attempts to answer natural language questions based on large-scale unstructured documents. Yet, the existing QA datasets are essentially constructed from Wikipedia or other synchronic document collections\footnote{Note that existing news datasets such as CNN/Daily Mail \cite{hermann2015teaching} and NewsQA \cite{trischler2016newsqa} are more suited to MRC tasks rather than to ODQA task due to the cloze question type or the ambiguity prevalent in their questions as we will discuss later.
In addition, their underlying document collections span relatively short time periods, which are also quite recent (such as after June 2007 or April 2010).}. 
The lack of large-scale datasets hinders the development of ODQA on document archives such as news article archives where Temporal IR \cite{campos2014survey, INR-043} techniques need to be utilized. Note that ODQA on historical document collections can be useful in many cases such as providing support for journalists who wish to relate their stories to certain past events, historians who investigate the past as well as employees of diverse professions, such as insurance or broad finance sectors, who wish to assess current risks based on historical accounts in order to support their decision making. As indicated in previous studies \cite{wang2020answering, wang2021improving}, synchronic document collections like Wikipedia cannot successfully answer many minor or detailed questions about the events from the past since the relevant data for answering those questions is only available in primary sources preserved in the form of large archival document collections. 

To overcome these shortcomings of existing QA datasets, we devise a novel framework that assists in the creation of a diverse, large-scale ODQA dataset over a temporal document collection. The framework utilizes automatic question generation as well as a series of carefully-designed filtering steps to remove poor quality instances. As an underlying archival document collection, we use the New York Times Annotated Corpus (NYT corpus) \cite{sandhaus2008new}, which contains over 1.8 million news articles published between January 1, 1987 and June 19, 2007. The NYT corpus has been frequently used over the recent years for many researches in temporal IR, temporal news content analysis, archival search, historical analysis and in other related tasks \cite{campos2014survey, INR-043}. 
The final dataset that we release, ArchivalQA, contains 532,444 data instances and is divided into different sub-parts based on question difficulty and the presence of temporal expressions.

We choose a semi-automatic way to construct our dataset for several reasons. First, manually generating questions would be too costly as it requires knowledge of history from annotators. Second, since question generation (QG) has recently attracted considerable attention, the available models already achieve quite good performance. Third, current ``data-hungry'' complex neural network models require larger and larger datasets to maintain high performance. Finally, synthetic datasets have been effective in boosting deep learning models' performance and are especially useful in use cases involving distant target domains with highly specialized content and terminology, for which there is only a small amount of labeled data \cite{walonoski2020synthea, li2020diverse, feng2020genaug}. 
We then approach the dataset generation based on a cascade of carefully designed filtering steps that remove low quality questions from a large initial pool of generated questions. 

To sum up, we make the following contributions in this work:
\begin{itemize}
    \item We propose one of the largest ODQA datasets for news collections\footnote{The largest existing dataset that uses news articles, CNN/Daily Mail dataset \cite{hermann2015teaching}, has been created based on a straightforward cloze test and thus cannot be considered as a proper ODQA dataset.}, which is not only spanning the longest time period compared to other QA datasets, but it also provides detailed questions on the events that occurred from 14 to 34 years ago. 
    \item We propose an approach to generate large datasets in an inexpensive way, whose resulting questions tend to be non-ambiguous and of good quality, thus having only a single potential answer. Compared with other QG methods, 
    most questions generated by our approach are clear and non-ambiguous, and thus they can be especially useful in improving computational approaches to education, e.g., to support generating questions for exams.
    \item We undertake comprehensive analysis of the generated dataset, which does not only show the quality and utility of the resulting data, but also proves the effectiveness of our QG framework.
\end{itemize}

\section{Related Work}
\begin{table*}
\footnotesize
  \caption{Comparison of related datasets. Note that there are more synchronic datasets that are not listed here \cite{zhu2021retrieving} (roughly about 30 common QA datasets based on our investigation).}
  \vspace{-0.5em}
  \label{tab_compareddataset}
  \renewcommand{\arraystretch}{1.2}
  \setlength{\tabcolsep}{0.3em}
  \centering
  \begin{tabular}{llllllc}
  \hline
  \textbf{Dataset} & \textbf{\#Questions} & \textbf{Answer Type} & \textbf{Question Source} & \textbf{Corpus Source} & \textbf{Synch/Diach} & \textbf{Non-ambiguous}\\ \hline
  \vspace{-0.1em}
  \makecell[l]{MS MARCO \cite{nguyen2016ms}} & 1M &  \makecell[l]{Generative, \\ Boolean} & Query logs  & Web documents & Synchronic & \xmark\\ \hline
  \vspace{-0.1em}
  \makecell[l]{SQuAD 1.1 \cite{rajpurkar2016squad}} & 108K & Extractive  & Crowd-sourced & Wikipedia & Synchronic & \xmark\\ \hline
  \vspace{-0.1em}
  \makecell[l]{SQuAD 2.0 \cite{rajpurkar2018know} }& 158K & Extractive  & Crowd-sourced & Wikipedia &  Synchronic & \xmark\\ \hline
  \vspace{-0.1em}
  \makecell[l]{NaturalQuestions \cite{kwiatkowski2019natural}} & 323K &  \makecell[l]{Extractive, \\ Boolean}  & Query logs & Wikipedia & Synchronic & \xmark\\ \hline
  \vspace{-0.1em}
  \makecell[l]{CNN/Daily Mail \cite{nallapati2016abstractive}} & 1M & Cloze  & \makecell[l]{Automatically \\ Generated} & News & \makecell[l]{Diachronic \\ (2007/04 - 2015/04)} & \xmark \\ \hline
  \vspace{-0.1em}
  \makecell[l]{NewsQuizQA \cite{lelkes2021quiz}} & 20K & Multiple-choice  & Crowd-sourced & News & \makecell[l]{Diachronic \\ (2018/06-2020/06)} & \xmark\\ \hline
  \vspace{-0.1em}
  \makecell[l]{NewsQA \cite{trischler2016newsqa}} & 119K & Extractive  & Crowd-sourced & News & \makecell[l]{Diachronic \\ (2007/04-2015/04)} & \xmark\\ \hline
  \vspace{-0.1em}
  ArchivalQA & 532K & Extractive & \makecell[l]{Automatically \\ Generated} & News & \makecell[l]{Diachronic \\ (1987/01-2007/06)} & \cmark\\
  \hline
\end{tabular}
\end{table*}


\subsection{QA Benchmarks}
In the recent years, a large number of QA benchmarks have been introduced \cite{zeng2020survey, baradaran2020survey, dzendzik2021english, rogers2021qa}.
The SQuAD 1.1 \cite{rajpurkar2016squad} consists of question-answer pairs that are made from the paragraphs of 536 Wikipedia articles. This dataset was later extended by SQuAD 2.0 \cite{rajpurkar2018know} that contains also unanswerable questions. 
NarrativeQA \cite{kovcisky2018narrativeqa} uses a different resource, the summaries of movie scripts and books, to create its question-answer pairs. 
MS MARCO \cite{nguyen2016ms} and NaturalQuestions \cite{kwiatkowski2019natural} use the search query logs of Bing and Google search engines as the questions, and the retrieved web documents and Wikipedia pages are collected as the evidence documents. 

Most of the existing datasets are designed over synchronic document collections, such as books, Wikipedia articles and web search results. While there are some MRC datasets created based on the news collections, they mostly belong to the cloze style datasets, such as CNN/Daily Mail \cite{nallapati2016abstractive}, WhoDidWhat \cite{onishi2016did} and ReCoRD \cite{zhang2018record}, with the aim to predict the missing word in a passage rather than to answer proper questions; hence these datasets cannot be used in the ODQA task.
Although \citet{lelkes2021quiz} constructed the NewsQuizQA dataset based on news articles, too, its questions belong to the multiple-choice type, which are easier to be answered, and the dataset contains only 20K question-answer pairs. The question-answer pairs were also obtained from only 5K summaries derived from the recent news articles. In addition, NewsQuizQA has been designed as a dataset for generating the quiz-style question-answer pairs.

To the best of our knowledge, NewsQA \cite{trischler2016newsqa} is the only MRC dataset in which an answer is a text span which is created based on the temporal document collection, the CNN news articles. 
However, our dataset has significant differences when compared to NewsQA. First, dataset size of NewsQA is much smaller than ours (119K vs. 532K). Second, its underlying CNN corpus contains less news articles which span shorter and also more recent time period (93k articles from 2007/04 to 2015/04 vs. 1.8M articles from 1987/01 to 2007/06 as in our case). We have also found that NewsQA is essentially appropriate for the MRC task and is not very suitable for the ODQA task. This is because many questions require additional background knowledge about their original paragraphs for understanding and correctly answering them. 
These questions tend to be ambiguous, unclear and generally impossible to be answered over the large news collection, because they are not specific enough and tend to have multiple correct answers (e.g., the questions \textit{“When were the findings published?”}, \textit{“Who drew inspiration from presidents?”} and \textit{“Whose mother is moving to the White House?”}\footnote{These questions are actually shown as examples on the NewsQA website: \url{https://www.microsoft.com/en-us/research/project/newsqa-dataset/stats/}}). Note that questions on some QA datasets also have similar characteristics, for example, \citet{min2020ambigqa} found that over half of the questions in the NaturalQuestions are ambiguous, with diverse sources of ambiguity such as event and entity references.
Finally, the questions in NewsQA have been created from 7 times less articles than in our final dataset (12,744 vs. 88,431). 

Thus, the goal of this work is to create a large-scale, non-ambiguous QA dataset over a long-term historical document collection that can promote the development of ODQA systems on historical news archives. In Tab.~\ref{tab_compareddataset} we summarize differences between ArchivalQA and the most related datasets. 

\subsection{Automatic Question Generation}
In the recent years, automatic question generation (AQG) has greatly advanced thanks to deep learning techniques, and it has received increasing attention due to its wide applications in education \cite{kurdi2020systematic}, dialogue systems \cite{wang2018learning}, and question answering \cite{duan2017question}. 
Diverse types of neural sequence-to-sequence models have been proposed for the AQG task. \citet{zhao2018paragraph} introduce the model incorporating paragraph-level inputs - the first model that achieved large improvement over sentence-level inputs. \citet{sun2018answer, kim2019improving} improved the performance by encoding answer positions, which can help to generate better-quality answer-focused questions. Some works also propose QG models under particular constraints, e.g., controlling the difficulty \cite{gao2018difficulty} and topic \cite{hu2018aspect} of the generated questions. In addition, models that can jointly learn to ask (QG) as well as answer questions (QA) have been also proposed \cite{wang2017joint, sachan2018self}. 
Moreover, it has been shown that having a large, even synthetic dataset, is useful for training QA models with different objectives. 
For example, \citet{puri2020training} train their model using only the synthetic data and obtain state-of-the-art performance on SQuAD dev set.
\citet{shakeri2020end} improve the performance of models in target domains by utilizing the synthetic dataset. 
\citet{saxena2021question} demonstrate that the large size  model-generated dataset can help in training temporal reasoning models.
\citet{lewis2019unsupervised} propose to use unsupervised question generation (e.g., template/rule-based methods) to tackle unsupervised QA task, a setting in which no aligned question, neither context no answer data are available. They demonstrate that their method can outperform early supervised models on SQuAD 1.1 without using the SQuAD training data, and modern QA models can learn to answer human questions surprisingly well using only synthetic training data.
In addition, some existing Visual Question Answering (VQA) datasets, such as COCO-QA \cite{ren2015exploring} and Visual Madlibs \cite{yu2015visual}, have also had AQG applied to generate their questions.

However, we argue that most of the questions automatically generated by the above models can be applied only to machine reading comprehension setting when a relevant paragraph is given. When used for ODQA task, some questions turn to be ambiguous and result in several potential answers (the same problem we observed in the NewsQA dataset as discussed above). Therefore, we propose a semi-automatic method that combines AQG with a cascade of customized filtering steps to generate the final dataset, whose resulting questions are non-ambigous and of good quality. We believe that this approach could be also applied to other types of temporal collections.
Such framework would be also useful in education field, where forming good and clear questions is crucial for evaluating students knowledge and for stimulating self-learning.


\section{Methodology}
\begin{figure*}[t]
  \centering \includegraphics[width = 0.8\textwidth]{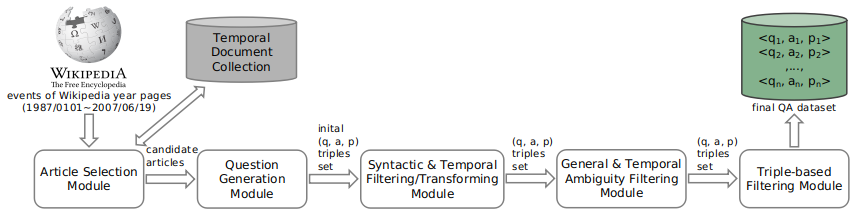}
  \caption{Dataset generation framework}
  \label{fig_framework}
   \vspace{-0.4em}
\end{figure*}

We introduce here the framework that generates and selects questions from temporal document collections. Fig.~\ref{fig_framework} shows its architecture which consists of five modules: \textit{Article Selection Module}, \textit{Question Generation Module}, \textit{Syntactic \& Temporal Filtering/Transforming Module}, \textit{General \& Temporal Ambiguity Filtering Module} and \textit{Triple-based Filtering Module}. All these modules are described below.

\subsection{Article Selection Module}
This module is responsible for deciding which articles are used to generate the initial set of questions. We use two approaches for selecting the articles.

\subsubsection{Selection based on Wikipedia Events}
The first one relies on the short descriptions of important events available in Wikipedia year pages\footnote{List of year pages: \url{https://en.wikipedia.org/wiki/List_of_years} and events for an example year: \url{https://en.wikipedia.org/wiki/1989}} as the seeds to find related articles. Since we utilize the NYT corpus, we use 2,976 event descriptions which occurred between January 1, 1987 and June 19, 2007. Then, for each such event description, we select keywords to be used as search queries for retrieving articles related to this description from the news archive. 
We choose Yake!\footnote{Yake! is available in the PKE tookit: \url{https://github.com/boudinfl/pke}} \cite{campos2020yake} as our keyword extraction method, which is a state-of-the-art unsupervised approach that relies on statistical features to select the most important keywords. Next, the query composed of the extracted keywords is sent to the ElasticSearch\footnote{\url{https://www.elastic.co/}} installation which returns the top 25 relevant documents ranked by BM25. Finally, 53,991 news articles are obtained in this way to be used for generating questions.

\subsubsection{Random Selection}
The second way is to randomly select long news articles from the corpus, which have at least 100 tokens. Based on this step, additional 55,000 news articles are collected.

We followed these two ways because we wanted the final dataset to contain questions related to important past events as well as also questions on minor issues, especially ones which are likely not recorded in Wikipedia, and thus more challenging and unique\footnote{In the experiments we actually show that only a small number of our questions can be successfully answered when using Wikipedia.}. 

\subsection{Question Generation Module}
The second step is to generate questions from the collected articles. We first separate articles into paragraphs and then use a neural network model to generate candidate questions from each paragraph. 
We apply T5-base \cite{raffel2019exploring} - a recent, large, pre-trained Transformer encoder-decoder model.
We note that, same as us, \citet{lelkes2021quiz} have used QG methods to generate questions from news articles in an automatic way, although in their case PEGASUS model was utilized to generate the questions using the NewsQuizQA dataset. However, we did not choose PEGASUS-base model
since we found that it generates questions which sometimes contain information not found in the underlying documents (probably due to the Gap Sentences Generation pre-training task that the PEGASUS-base model applies).
Furthermore, the questions generated by \citet{lelkes2021quiz} belong to the quiz-style multiple-choice type which is not suitable for ODQA. 

We fine-tune our model using SQuAD 1.1\footnote{We decided not to use NewsQA for training as it contains too many ambiguous questions.} \cite{rajpurkar2016squad} whose inputs are the answers together with their corresponding paragraphs, and the questions form the outputs. The final model achieves good performance on the SQuAD 1.1 dev set (the scores of BLEU-4, METEOR, ROUGE-L are 21.19, 26.48, 42.79, respectively).
After fine-tuning the model, every named entity\footnote{We use the named entity recognizer from spaCy: \url{https://github.com/explosion/spaCy}.} in a given paragraph of each article is labeled as an answer, and is used along with the paragraph as the input to the model. 
Note that the answers of many QA datasets, such as CNN/Daily Mail \cite{nallapati2016abstractive}, TriviaQA \cite{joshi2017triviaqa}, Quasar-T \cite{dhingra2017quasar}, SearchQA \cite{dunn2017searchqa} and XQA \cite{liu2019xqa}, are also mainly in the form of entities (e.g., 92.85\% of the answers in TriviaQA are Wikipedia entities), as this improves answering accuracy.
In addition, we restrict the number of tokens of the paragraphs and of the corresponding sentences which include the answers. More specifically, the paragraphs that have less than 30 tokens are eliminated. Additionally, the answers whose corresponding sentences have less than 10 tokens are discarded.
Finally, we generated 6,408,036 questions in this way from 1,194,730 paragraphs of 106,197 news articles. 

\subsection{Syntactic \& Temporal Filtering/Transforming Module}
This module consists of 8 basic processing steps that further remove or transform the candidate question-answer pairs obtained so far:
\begin{enumerate}
  \item Remove questions that do not end with a question mark (107,586 such questions removed). 
  \item Remove questions whose answers are explicitly indicated inside the questions' content (127,212 questions removed). For example, question like \textit{"Where did Mr. Roche serve in Vietnam?"} that has gold answer "Vietnam" is removed.
  \item Remove duplicate questions. The same questions generated from different paragraphs are removed (492,257 questions removed).
  \item Remove questions that have too few or too many named entities. Questions without any named entity or with more than 7 named entities are eliminated (1,310,621 questions removed).
  \item Remove questions that are too short or too long. Questions that contain less than 8 or more than 30 tokens are dropped (463,726 questions removed).
  \item Remove questions with unclear pronouns, for example, “\textit{What was the name of the agency that she worked for in the Agriculture Department?}” (63,300 questions removed). The details of this step are described in Appendix~\ref{sec:appendix_1}.
  \item Transform relative temporal information in questions to absolute temporal information. 
  For example, \textit{“How many votes did President Clinton have in New Jersey last year?”} is transformed to \textit{“How many votes did President Clinton have in New Jersey in 1996?”} (140,658 questions transformed). The details are given in Appendix~\ref{sec:appendix_2}.
  \item Transform relative temporal information of the answers of generated questions to absolute temporal information. We apply the same approach as in the previous step. For example, the answers to questions \textit{“When did Rabbi Riskin write about protests by West Bank settlers in Israel?”} and \textit{“When were the three teenagers convicted of murdering Patrick Daly?”}, which are “Aug. 7” and “yesterday”, respectively, are transformed to “August 07, 1995” and “June 15, 1993”, by incorporating the articles' publication dates: `1995-08-12' and `1993/06/16' (279,671 answers transformed in this way).
  \end{enumerate}
  


\subsection{General \& Temporal Ambiguity Filtering Module}
\subsubsection{Filtering by Content Specificity}
Sentence specificity is often pragmatically defined as the level of detail of the information contained in the sentence \cite{louis2011automatic, li2015fast}. In contrast to specific sentences that contain informative messages, general sentences do not reveal much specific information (e.g., overview statements). In the examples shown below, the first sentence is general as it is clearly less informative than the second sentence (specific one), and is not suitable to be used for question generation.
\begin{enumerate}[label=\arabic*)]
\item \textit{"Despite recent declines in yields, investors continue to pour cash into money funds."}
\item \textit{"Assets of the 400 taxable funds grew by \$1.5 billion during the last week, to \$352.7 billion."}
\end{enumerate}
Thus, in this step, we aim to remove questions that have been generated from general sentences.
We use the training dataset from \citet{ko2019domain}, which is composed of three publicly available, labeled datasets \cite{louis2012corpus,li2015fast,li2016improving}. The combined dataset contains 4,342 sentences taken from news articles together with their sentence-level binary labels (general vs. specific). We partition this dataset randomly into the training set (90\%), and the test set (10\%). We next fine-tune three Transformer-based classifiers: BERT-based model \cite{devlin2018bert}, RoBERTa-base model \cite{liu2019roberta} and ALBERT-base model \cite{lan2019albert}, such that each classifier consists of the corresponding pre-trained language model followed by a dropout layer and a fully connected layer.
We finally choose RoBERTa-base model \cite{liu2019roberta} as our specificity-determining model because it achieves the best results on the test set - 84.49\% accuracy. Finally, we discard all questions whose underlying sentences 
from which they were generated
have been classified by the above-described approach as general. This filtering step removed 952,398 questions. Few examples of the removed questions are given in Tab.~\ref{tab_removed_questions} in Appendix.
\subsubsection{Filtering by Temporally Ambiguity}
\begin{table}
\footnotesize
  \caption{
  Temporal ambiguity of example questions.}
  \vspace{-0.5em}
  \label{tab_tempambi}
  \renewcommand{\arraystretch}{1.6}
  \setlength{\tabcolsep}{0.21em}
  \centering
  \begin{tabular}{|P{0.03\textwidth}|P{0.26\textwidth}|P{0.10\textwidth}|}
  \hline
  \textbf{No.} &
  \textbf{Question} & \textbf{Ambiguity} \\
    \hline
    1  & \makecell[c]{Who did President Bush announce he would \\ submit a trade agreement with?}  &  \makecell[c]{Temporally\\ ambiguous}\\
    \hline
    2  &  \makecell[c]{When was the National Playwrights\\ Conference held?}  &  \makecell[c]{Temporally \\ambiguous}\\
    \hline
    3  &  \makecell[c]{Who won the Serbian presidential election \\in October, 2002?}  &  \makecell[c]{Temporally\\ non-ambiguous}\\
    \hline
    4  &  \makecell[c]{Where did the Tutsi tribe massacre \\thousands of Hutu tribesmen?}  &  \makecell[c]{Temporally\\ non-ambiguous}\\
    \hline
\end{tabular}
\end{table}

When manually analyzing the resulting dataset we have observed that some questions are problematic due to their temporal ambiguity, e.g., \textit{“How many people were killed by a car bomb in Baghdad?”}. Such questions can be matched to several distinct events. The first and the second generated example questions in Tab.~\ref{tab_tempambi} exhibit such characteristics; the correct answers of such questions should be actually a list of answers rather than a single answer. 
However, the datasets having multiple correct answers for each question are quite rare in the current ODQA field \cite{zhu2021retrieving} (we are only aware of AMBIGQA dataset \cite{min2020ambigqa} which contains multiple possible answers to ambiguous questions). This might be because it would not be clear how to rank systems as some of the ground-truth answers might be more preferred than others. In our case, for example, some events related to the ambiguous questions could be more important or more popular than other related events.
Also, and perhaps more importantly, finding all the correct answers to such questions is quite difficult, if not impossible, within a large news collection (especially an archival one that spans two decades such as ours). Hence, we decided to remove temporally ambiguous questions, however we will make them available for the community to download as a separate data, should anyone be interested in studying questions of this type.  

We define temporally ambiguous questions as ones that have multiple correct and different answers over time.
Note that temporally ambiguous questions are specific to temporal datasets like ours, and consequently they have not been studied before.
Since there is no readily available dataset for detecting temporally ambiguous questions, we have manually labeled 5,500 questions obtained from the previous filtering steps\footnote{ This dataset will be also made freely available, as it could be useful for improving QG research.}.
Then, we again fine-tuned three Transformer-based classifiers, same as when training the specificity-evaluating model. The BERT-based model \cite{devlin2018bert} has been finally chosen as it performs best on the test set achieving 81.82\% accuracy. We then used it to remove 1,823,880 questions classified as temporally ambiguous\footnote{As mentioned before, we will also release the data of temporally ambiguous questions, which could be useful for developing systems that can provide multiple possible answers.}. Similarly, in Tab.~\ref{tab_removed_questions} in Appendix, we also give few examples of the removed ambiguous questions.

\begin{figure}
\vspace{-.6em}
  \centering \includegraphics[width = 0.48\textwidth]{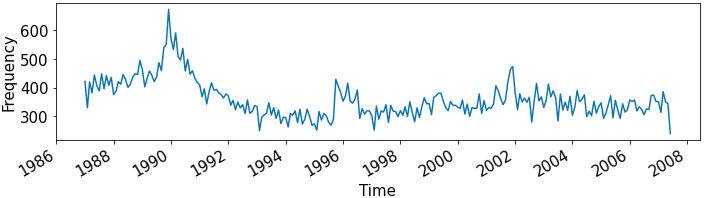}
  \vspace{-1.5em}
  \caption{Distribution of articles used in ArchivalQA}
  \label{fig_docdistri}
  \vspace{-1.1em}
\end{figure}

\subsection{Triple-based Filtering Module}
In the final module, we aim to remove remaining poor quality data instances by analyzing the entire <question, answer, paragraph> triples. Some instances are still problematic due to several reasons (e.g., questions with incorrect answers, questions containing information not found in paragraphs, or other wrong questions that have not been filtered out by the previous filtering stages). To construct the last filter we first created a dedicated dataset by asking 10 annotators to label 10k samples selected from the results obtained after applying the previously-introduced filtering stages. The labels were either "Good" or "Bad" based on <paragraph, question, answer> triples\footnote{This dataset will be also available.}. The annotators had to not only consider the particular problems we discussed before, but also check whether the questions are grounded in their paragraphs and whether they can be answered by their answers, and whether the questions are grammatically correct or not. The dataset, that contains 5,699 "Good" questions and 4,301 "Bad" questions, was then randomly split into the training set (90\%), and the test set (10\%). Then, we trained a RoBERTa-base model \cite{liu2019roberta} that takes the triples as the input after adding a special token ([SEP]) to question-answer pair and paragraph of each sample. We set a high threshold that permits only the predicted good triples with probabilities higher than 0.99 be chosen as the final good triples. This last filtering step resulted in the precision of finding good triples to be 86.74\% on our test set. Finally, we removed 534,612 questions whose corresponding triples were classified as bad.

\section{Dataset Analysis}
\subsection{Data Statistics}
After all the above filtering steps, we have finally obtained the dataset which includes 532,444 question-answer pairs that were derived from 313,100 paragraphs of 88,431 news articles. About half of the questions (263,292) come from the randomly selected articles, and the other questions (269,152) are based on articles that were selected based on Wikipedia events. This provenance information is recorded for each question.
Paragraph IDs are also appended to each question-answer pair to let ODQA systems explicitly train their IR components. We partition the entire dataset randomly into the training set (80\%, 425,956 examples), the development set (10\%, 53,244 examples), and the test set (10\%, 53,244 examples). Tab.~\ref{tab_datasetexa} shows few examples. 
More detailed dataset statistics are presented in Tab.~\ref{dataset_sta}. Fig.~\ref{fig_docdistri} shows also the temporal distribution of documents used for producing ArchivalQA questions. 

\begin{figure}
  \centering \includegraphics[width=0.48\textwidth,height=43mm]{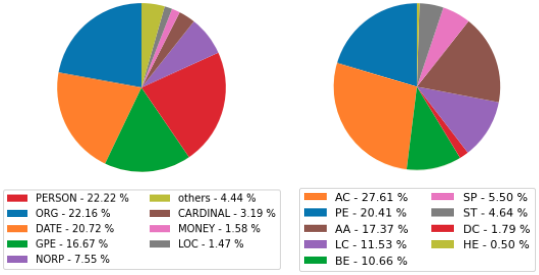}
  \vspace{-5.7mm}
  \caption{Left: Answers' named entity distribution (``others'': named entities that account for a very small part (< 1\%)). Right: Questions' category distribution (``AC'': ``arts \& culture'', ``PE'': ``politics \& elections'', ``AA'': ``armed conflicts \& attacks'', ``LC'': ``law and crime'', ``BE'': ``business \& economy'', ``SP'': ``sport'', ``ST'': ``science \& technology'', ``DC'': ``disasters \& accidents'', ``HE'': ``health \& environment'').
  }
  \label{fig_ansent}
  \vspace{-0.5em}
\end{figure}

\begin{figure}
  \centering \includegraphics[width=0.37\textwidth,height=0.37\textwidth]{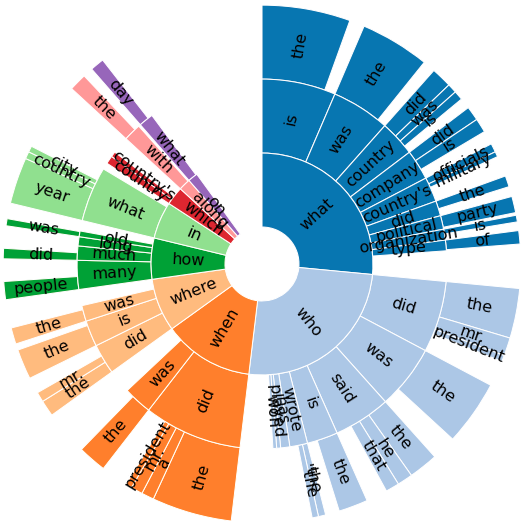}
  \caption{Trigram prefixes of ArchivalQA questions}
  \label{fig_tridis}
  \vspace{-0.9em}
\end{figure}

We have also analyzed the named entity types\footnote{18 entity types used by NE recognizer in spaCy.} of the answers in the dataset.
As shown in the left pie chart in Fig.~\ref{fig_ansent}, the answers that belong to PERSON, ORG, DATE, GPE and NORP\footnote{NORP denotes nationality or religious or political groups; for example, "Catholic".} account for a large part of ArchivalQA. Further, the right hand side's pie chart in Fig.~\ref{fig_ansent} shows the distribution of 9 event categories of the questions that are classified by another dedicated classifier prepared by us, which has been trained based on the event dataset created by \citet{sumikawa2018system} achieving 85.86\% accuracy. We can see that ArchivalQA contains questions related to diverse event categories, while the "arts \& culture", "politics \& elections", "armed conflicts \& attacks", "law and crime" and "business \& economy" events account for a large portion of questions.
Fig.~\ref{fig_tridis} presents also the distribution of frequent
trigram prefixes. While nearly half of SQuAD questions are "what" questions \cite{reddy2019coqa}, the distribution of ArchivalQA is more evenly spread across multiple question types.

\begin{table*}[t]
\footnotesize
  \caption{ArchivalQA Dataset Examples. org\_answer, answer\_start, trans\_que, trans\_ans, and source represent the original answer text, its start index in the document, flag indicating whether the question has been transformed, flag showing whether the answer has been transformed and the selection method of the document used for producing the question, respectively. para\_id contains concatenated information of the document ID (the metadata of each article in the NYT corpus) and the ith paragraph used to generate the question.}
  \label{tab_datasetexa}
  \renewcommand{\arraystretch}{1.2}
  \setlength{\tabcolsep}{0.35em}
  \centering
  \begin{tabular}{|c|c|c|c|c|c|c|c|c|}
  \hline
  \textbf{id} & \textbf{question} & \textbf{answer} & \textbf{org\_answer} & \textbf{answer\_start} & \textbf{para\_id} & \textbf{trans\_que} & \textbf{trans\_ans} & \textbf{source}\\
    \hline
    train\_0 & \makecell[c]{Who claimed responsibility for the\\ bombing of Bab Ezzouar?} & Al Qaeda & Al Qaeda  & 184 & 1839755\_20 & 0 & 0 & wiki \\
    \hline
    train\_4 &  \makecell[c]{When did Tenneco announce it was\\ planning to sell its oil and gas operations?} & \makecell[c]{May 26, 1988} & today  & 103 & 148748\_0 & 0 & 1 & rand \\
    \hline
    val\_45 &  \makecell[c]{What threat prompted Mr. Paik's family \\to flee to Hong Kong?} & \makecell[c]{the Korean War} & the Korean War  & 327 & 1736040\_7 & 0 & 0 & wiki \\
    \hline
    test\_84 &  \makecell[c]{Along with the French Open, what other\\ tournament did Haarhuis win in 1998?} & \makecell[c]{Wimbledon} & Wimbledon  & 527 & 1043631\_15 & 1 & 0 & rand \\
    \hline
\end{tabular}
\end{table*}

\begin{table}
\footnotesize
  \caption{Basic statistics of ArchivalQA}
  \vspace{-0.6em}
  \label{dataset_sta}
  \renewcommand{\arraystretch}{1.2}
  \setlength{\tabcolsep}{0.35em}
  \centering
  \begin{tabular}{l|c}
  \hline
    \vspace{-0.2em}
    Number of QA pairs  &  532,444\\
    \vspace{-0.2em}
    Number of transformed questions  &  29,696\\
    \vspace{-0.2em}
    Number of transformed answers  &  47,972\\
    \hline
    \vspace{-0.2em}
    Avg. question length (words) &  12.43\\
    \vspace{-0.2em}
    Avg. questions / document &  6.02\\
    \vspace{-0.1em}
    Avg. questions / paragraph &  1.70\\
    \hline
\end{tabular}
\end{table}

\subsection{Model Performance}
\begin{table}[t]
\footnotesize
\centering
  \caption{Models' performance on ArchivalQA}
  \vspace{-0.5em}
  \label{tab_datasetperformance}
  \renewcommand{\arraystretch}{1.2}
  \setlength{\tabcolsep}{0.3em}
  \begin{tabular}{|p{0.19\textwidth}|P{0.06\textwidth}|P{0.06\textwidth}|}
  \hline
  \textbf{Model} & \textbf{EM} & \textbf{F1}  \\
    \hline
    \makecell[l]{DrQA-Wiki \cite{chen2017reading}} & 7.53 & 11.64\\
    \hline
    \makecell[l]{DrQA-NYT \cite{chen2017reading}}  &  38.13 &  46.12\\
    \hline
    \makecell[l]{DrQA-NYT-TempRes \cite{chen2017reading}} &  44.84 &  53.06\\
    \hline
    \makecell[l]{BERTserini-Wiki \cite{yang2019end}} & 10.19 & 16.25\\
    \hline
    \makecell[l]{BERTserini-NYT \cite{yang2019end}} &  54.30 & 66.05\\
    \hline
    \makecell[l]{BERTserini-NYT-TempRes \cite{yang2019end}} &  \textbf{56.34} & \textbf{68.93}\\
    \hline
    \makecell[l]{DPR-NYT \cite{karpukhin2020dense}} & 47.78 & 60.78\\
    \hline
    \makecell[l]{DPR-NYT-TempRes \cite{karpukhin2020dense}} & 52.93 & 64.98\\
    \hline
\end{tabular}
\end{table}

We use the following well-established ODQA approaches to show their results on ArchivalQA:
\begin{enumerate}
\item DrQA-Wiki \cite{chen2017reading}: DrQA combines a search component based on bigram hashing and TF-IDF matching with a multi-layer recurrent neural network model trained to extract answers from articles. We first test the DrQA model which uses Wikipedia as the knowledge source (DrQA's default knowledge source). With this setting we would like to test if Wikipedia alone could be sufficient for answering questions about the historical events.  
\item DrQA-NYT \cite{chen2017reading}: DrQA model which uses NYT.
\item DrQA-NYT-TempRes \cite{chen2017reading}: DrQA model which uses NYT archive and transforms the answers with relative temporal information by an approach similar to the one we used for transforming relative temporal information in Syntactic \& Temporal Filtering/Transforming Module (see the 7th and 8th steps of Sec. 3.3).
\item BERTserini-Wiki \cite{yang2019end}: BERTserini tackles end-to-end question answering by combining BERT \cite{devlin2018bert} with Anserini \cite{yang2017anserini} IR toolkit, with BM25 as the ranking function. We also first test BERTserini model using Wikipedia (BERTserini's default knowledge source).
\item BERTserini-NYT \cite{yang2019end}: BERTserini model which uses NYT.
\item BERTserini-NYT-TempRes \cite{yang2019end}: BERTserini model which uses NYT archive and transforms the relative temporal answers.
\item DPR-NYT \cite{karpukhin2020dense}\footnote{We have not decided to test DPR using Wikipedia as the knowledge source, due to considerable time cost required.}: Unlike previous ODQA approaches, this end-to-end QA model incorporates BERT \cite{devlin2018bert} reader module\footnote{The same reader module that is used in BERTserini model.} with dense retriever module that has been trained for 15 epochs using ArchivalQA dataset and NYT corpus. In the retriever module, the paragraphs and questions are represented by dense vector representations, computed using two BERT networks. The ranking function is given by the dot product between the query and passage representations.
\item DPR-NYT-TempRes \cite{karpukhin2020dense}: DPR model which uses NYT archive and transforms the relative temporal answers.
\end{enumerate}

We measure the performance of the above-listed models using exact match (EM) and F1 score - the two standard measures commonly used in QA research. The results of all the models are given in Tab.~\ref{tab_datasetperformance}. 
Firstly, we can observe that the models that utilize Wikipedia as the knowledge source perform much worse than the models that use NYT corpus, which is due to many questions being
about minor things or events that Wikipedia does not seem to record (or it describes them only shallowly). Secondly, the models that resolve implicit temporal answers perform better than the ones without this step. Temporal information resolution is then clearly important. Thirdly, 
we notice that BERTserini models outperform DrQA models by large margins. There are two possible reasons, one is that DrQA models retrieve the entire long articles containing many non-relevant sentences rather than short paragraphs; the other is that DrQA uses RNN-base reader component rather than a better choice which would be the BERT-base reader component. Finally, DPR models which use dense vector representations for retrieval also achieve relatively good results on both metrics.
Future work on combining dense retrieval with sparse retrieval could be studied to further improve the performance.

\subsection{Human Evaluation}
\begin{table}[t]
\footnotesize
  \caption{Human evaluation results of ArchivalQA}
  \vspace{-0.5em}
  \label{tab_he}
  \renewcommand{\arraystretch}{1.2}
  \setlength{\tabcolsep}{0.22em}
  \begin{tabular}{|P{0.08\textwidth}|P{0.11\textwidth}|P{0.09\textwidth}|P{0.11\textwidth}|}
  \hline
   \textbf{Fluency} & \textbf{Answerability} & \textbf{Relevance} & \textbf{Non-ambiguity}  \\
    \hline
    4.80 & 4.57 & 4.79 & 4.60 \\
    \hline
\end{tabular}
\vspace{-1.2em}
\end{table}

We finally conduct human evaluation on ArchivalQA to study the quality of the generated questions. We randomly sampled 5K question-answer pairs along with their original paragraphs and publication dates and asked 10 graduate students for their evaluation. The evaluators were requested to rate the generated questions from 1 (very bad) to 5 (very good) on four criteria: \emph{Fluency} measures if a question is grammatically correct and is fluent to read. \emph{Answerability} indicates if a question can be answered by the given answer. \emph{Relevance} measures whether a question is grounded in the given passage, while \emph{Non-ambiguity} defines if a question is non-ambiguous.
The average scores for each evaluation metric are shown in Tab.~\ref{tab_he}. Our model achieves high performance over all the metrics, especially on \emph{Fluency} and \emph{Relevance}. In addition, the \emph{Non-ambiguity} result is high, indicating that large majority of the questions are non-ambiguous.

We then examine the effectiveness of General \& Temporal Ambiguity Filtering Module by analyzing reasons as for why 10 annotators labelled 10k data samples as "Bad" for the Triple-based Filtering Module.
As shown in Tab.~\ref{10kquestions}, among 10k questions, there are 390 (3.90\%) questions labelled as "Bad" due to specificity problems, and 806 (8.06\%) questions have temporal ambiguity problems\footnote{Other "Bad" questions are the questions with incorrect answers, questions containing information not found in paragraphs, or questions with bad grammar, etc.)}. This relatively small numbers suggest that the General \& Temporal Ambiguity Filtering Module should have removed most of the questions with specificity or ambiguity issues. The final filtering step using the Triple-based Filtering Module is supposed to remove the remaining "Bad" questions by analyzing <question, answer, paragraph> at the same time.

\begin{table}
\footnotesize
  \caption{Statistics of the dataset used in Triple-based Filtering}
  \vspace{-0.6em}
  \label{10kquestions}
  \renewcommand{\arraystretch}{1.2}
  \setlength{\tabcolsep}{0.35em}
  \centering
  \begin{tabular}{l|c}
  \hline
    \vspace{-0.2em}
    Questions generated from general sentences  &  390\\
    \vspace{-0.2em}
    Temporally ambiguous questions & 806\\
    \vspace{-0.2em}
    Other "Bad" questions &  3,105\\
    \vspace{-0.2em}
    "Good" questions&  5,699\\
    \hline
    Total questions&  10,000\\
    \hline
\end{tabular}
\end{table}

\section{Sub-Dataset Creation}
We also distinguish subparts of the dataset based on the question difficulty levels and the containment of temporal expressions, which we believe could be used for training/testing ODQA systems with diverse strengths and abilities. Tab.~\ref{tab_subdataset} shows few randomly sampled examples for each of the four subdivisions of our dataset which we describe below.

\begin{table}[t]
\footnotesize
  \caption{ArchivalQA Sub-Dataset Examples}
  \label{tab_subdataset}
  \renewcommand{\arraystretch}{1.2}
  \setlength{\tabcolsep}{0.35em}
  \centering
  \begin{tabular}{|c|c|c|c|}
  \hline
  \textbf{id} & \textbf{question} & \textbf{answer} & \textbf{sub-dataset}\\
    \hline
    train\_134512 & \makecell[c]{What political party was Larry \\ Rockefeller a candidate for?} & Republican & Easy \\
    \hline
    val\_45168 &  \makecell[c]{What country did President Bush \\ send 30,000 troops to?} & Somalia & Difficult \\
    \hline
    train\_123981 &  \makecell[c]{What company was formed in 1986 by \\ the merger of Burroughs and Sperry?} & Unisys & Exp-Temp \\
    \hline
    test\_26021 &  \makecell[c]{What Prince was overthrown by Lon Nol?} & Sihanouk & Imp-Temp \\
    \hline
\end{tabular}
\end{table}

\begin{table*}[t]
\begin{center}
\footnotesize
  \centering \caption{Performance of different models over different Sub-Datasets}
  \label{subdataset_performance}
  \renewcommand{\arraystretch}{1.0}
  \centering
  \begin{tabular}{|m{32mm}|P{8mm}|P{8mm}|P{8mm}|P{9mm}|P{8mm}|P{8mm}|P{9mm}|P{9mm}|}
    \hline
    \multirow{2}{0.75cm}{\textbf{Model}} & \multicolumn{2}{c|}{\textbf{ArchivalQAEasy}} & \multicolumn{2}{c|}{\textbf{ArchivalQAHard}}& \multicolumn{2}{c|}{\textbf{ArchivalQATime}}& \multicolumn{2}{c|}{\textbf{ArchivalQANoTime}}\\
    \cline{2-9}
    & \textbf{EM} & \textbf{F1} & \textbf{EM} & \textbf{F1} & \textbf{EM} & \textbf{F1} & \textbf{EM} & \textbf{F1}\\
    \hline
DrQA-NYT \cite{chen2017reading} &42.10 & 51.97 & 22.81 & 31.24 & 31.32 & 42.17 &39.59 & 47.18\\ \hline
DrQA-NYT-TempRes \cite{chen2017reading} &48.41 & 57.26 & 27.37 & 34.02 & 33.19 & 44.01 &46.39 & 54.91\\ \hline
BERTserini-NYT \cite{yang2019end} &59.15 & 69.16 & 25.00 & 33.73 & 50.65 & 63.24 & 55.36 & 68.37\\ \hline
BERTserini-NYT-TempRes \cite{yang2019end} &\textbf{61.80} & \textbf{71.56} & 29.88 & 38.44 & \textbf{51.12} & \textbf{65.67} & \textbf{58.27} & \textbf{70.19}\\ \hline
DPR-NYT  \cite{karpukhin2020dense}&49.51 &63.56 & 44.38 & 52.81 &46.19 &58.35 &48.16 &61.38\\ \hline
DPR-NYT-TempRes  \cite{karpukhin2020dense}&55.50 &68.47 &\textbf{46.27} &\textbf{53.95} &46.87 &58.93 &54.27 &66.39\\ \hline
  \end{tabular}
  \end{center}
\end{table*}

\subsection{Difficult/Easy Questions Dataset}
We created two sub-datasets (called ArchivalQAEasy and ArchivalQAHard) based on the difficulty levels of their questions, such that 100,000 are easy and another 100,000 are difficult questions. 
We use open-source Anserini IR toolkit with BM25 as the ranking function to create these subsets. The samples are labeled as easy if the paragraphs used to generate the questions appeared within the top 10 retrieved documents; otherwise they are considered difficult. 
We then partitioned both these sub-datasets randomly into the training set (80\%, 80,000 examples), the development set (10\%, 10,000 examples), and the test set (10\%, 10,000 examples).

\subsection{Division based on Time Expressions}
We created the next two sub-datasets based on the temporal characteristics of their questions. In particular, we constructed two sub-datasets containing 75,000 questions with temporal expressions and 75,000 without temporal expressions (called ArchivalQATime and ArchivalQANoTime, respectively). 
We used SUTime \cite{chang2012sutime} combined with our handcrafted rules to collect the former questions, while the latter were randomly chosen questions without temporal expressions. Note that questions with temporal expressions should let ODQA systems limit the search time scope from the entire time frame of the news archive to the narrower time periods specified by the temporal expressions contained in these questions. For example, for the question \textit{"Which team won the 1990 World Series?"}, the answers could be just searched within documents published during (or perhaps also some time after) 1990. Same as with ArchivalQAEasy and ArchivalQAHard, both ArchivalQATime and ArchivalQANoTime were randomly split into the training (80\%, 60,000 examples), development (10\%, 7,500 examples), and test sets (10\%, 7,500 examples).

\subsection{Model Performance on Sub-Datasets}
Tab.~\ref{subdataset_performance} presents the performance of different models over the four sub-datasets discussed above. We can see that all the models achieve better results on ArchivalQAEasy than on ArchivalQAHard, indicating that the questions of ArchivalQAHard tend to be indeed harder to answer. For example, the improvement of BERTserini-NYT-TempRes is in the range of 106.83\% and 86.16\% on EM and F1 metrics, respectively. 
However, DPR models using dense vector representations for retrieving relevant paragraphs are subject to a small performance drop on two sub-datasets (ArchivalQAEasy and ArchivalQAHard) and they manage to surpass the other ODQA approaches that use sparse retrievers by large margins on ArchivalQAHard. For example, when considering DPR-NYT-TempRes model on ArchivalQAHard and ArchivalQAEasy, the improvements are only 19.95\% and 26.91\% on EM and F1, respectively. When comparing DPR-NYT-TempRes with BERTserini-NYT-TempRes on ArchivalQAHard, the improvements are 54.85\% and 40.35\% on EM and F1 metrics, respectively. This is likely because questions in ArchivalQAHard contain less lexical overlap with the NYT articles while DPR excels at semantic representation and handles lexical variations well.
When considering ArchivalQATime and ArchivalQANoTime, the models perform slightly better on ArchivalQANoTime.
A possible reason for that can be that such temporal signals are currently just used as usual textual information (rather than being utilized as time selectors) which can even cause harm, despite the fact that time expressions actually constitute an important feature. Future models should pay special attention to such temporal signals.

\section{Dataset Use}
Our dataset can be used in several ways. First, ODQA models can use the questions, answers and paragraphs\footnote{Note that another way to use the dataset is to train models without using the paragraph information \cite{lee2019latent}.} for training their IR and MRC modules \cite{karpukhin2020dense, ding2020rocketqa} on a novel kind of data that poses challenges in terms of highly changing contexts of different years, high temporal periodicity of events and rich temporal signals in terms of document timestamps and temporal expressions embedded in document content. As shown in \cite{wang2020answering, wang2021improving} systems that utilize such complex temporal signals (using Temporal IR approaches or others) achieve better results than conventional approaches. 

When it comes to the underlying news dataset, most systems would use our QA pairs against the NYT corpus. They might however potentially use other temporal news collections that temporally align with the NYT collection (i.e., ones that also span 1987-2007), although naturally this would make the task more challenging. It might be even feasible to consider answering our questions using synchronic knowledge bases such as Wikipedia, although as we have observed earlier, Wikipedia seems to lack a lot of detailed information on the past. The questions in our dataset are often specific and minor, and relate to relatively old events, hence they may be different than questions in other popular ODQA datasets. Such questions can be particularly valuable considering that the true utility of QA systems lies in answering hard questions that humans cannot (at least easily) answer by themselves.
Finally, system testing and comparison can be made to be more fine-grained based on the question difficulty and the occurrence of temporal components contained in questions. Also, another practical application could be to use our generated questions for education, e.g., for evaluating students knowledge and stimulating self-learning in history courses.
\vspace{-.8em}
\section{Conclusions}
We introduce in this paper a novel large-scale ODQA dataset for answering questions over a long-term archival news collection, with the objective to foster the research in the field of ODQA on news archives.
Our dataset is unique since it covers the the longest time period among all the ODQA datasets and deals with events that occurred in a relatively distant past.
An additional contribution is that we consider and mitigate the problem of temporally ambiguous questions for temporal document datasets. While this issue has not been observed in other ODQA datasets and researches, it is of high importance in long-term temporal datasets such as news archives.
Finally, we demonstrate a semi-automatic pipeline to generate large datasets via a series of carefully designed filtering steps.

\bibliographystyle{ACM-Reference-Format}
\bibliography{sample-base}

\appendix
\section{Appendix}
\label{sec:appendix}
\subsection{Unclear Pronouns Questions Removal}
\label{sec:appendix_1}
The questions with unclear pronouns are removed in the 6th step of the Syntactic \& Temporal Filtering/Transforming Module.
We first utilize part-of-speech tagger in spaCy to obtain the fine-grained POS information of each token in the generated questions. The questions whose tokens are classified as "PRP" or "PRP\$" are collected as the initial set of unclear-pronoun questions. Then we utilize the novel coreference resolution tool (NeuralCoref \cite{clark2016deep}) to obtain the coreference results of each sentence in the question set. For example, for the question "\textit{When did Sampras win his first Grand Slam?}", the information that 'his' points to 'Sampras' is derived. Then we apply several heuristic rules to collect only clear-pronoun questions.
A sentence is considered correct if its pronoun points to named entities appearing inside the question's content (e.g., 'Sampras' in the previous example), or if the question asks about the actual resolution of the pronoun (e.g., "\textit{Who dived into rough waters near her home in Maui to save a Japanese woman?}"), etc.
\subsection{Relative Temporal Information Transformation}
\label{sec:appendix_2}
The relative temporal information in questions and answers is transformed in the 7th and 8th step of the Syntactic \& Temporal Filtering/Transforming Module. We apply SUTime \cite{chang2012sutime} to recognize temporal expressions, and we use the publication date information of the articles, which include the paragraphs used to generate the question, as the reference date to transform the relative temporal information. Note that we do not transform all the temporal expressions in the entire corpus, since this would be too time-consuming. Additionally, this would change the original contents of the articles in the corpus, the situation which we try to avoid. Any systems that will use our dataset should see only the original, unchanged content of NYT's news articles for answering our dataset's questions. We expect that models which need to use temporal expressions should utilize article timestamps to resolve temporal expressions.


\begin{table}[h]
\footnotesize
  \caption{Examples of Questions Removed by the General \& Temporal Ambiguity Filtering Module.}
  \vspace{-0.5em}
  \label{tab_removed_questions}
  \renewcommand{\arraystretch}{1.6}
  \setlength{\tabcolsep}{0.21em}
  \centering
  \begin{tabular}{|P{0.03\textwidth}|P{0.26\textwidth}|P{0.08\textwidth}|P{0.08\textwidth}|}
  \hline
  \textbf{No.} &
  \textbf{Question} & \textbf{Answer}  & \textbf{Type} \\
    \hline
    1  &  \makecell[c]{Who goes to Central Park to walk, touch\\  grass, play?}  &  \makecell[c]{New Yorkers} &  \makecell[c]{General}\\
    \hline
    2  &  \makecell[c]{The Italian economy has been deteriorating \\ compared to what other country?}  &  \makecell[c]{Germany}  &  \makecell[c]{General}\\
    \hline
    3  &  \makecell[c]{Who is a nice, sweet Southern boy that\\ people underestimate?}  &  \makecell[c]{Bobby}  &  \makecell[c]{General}\\
    \hline
    4  & \makecell[c]{How many countries are in the World \\ Trade Organization?}    &  \makecell[c]{142}  &  \makecell[c]{Temporally\\ ambiguous}\\
    \hline
    5  &  \makecell[c]{What country agreed to normalize relations \\with the United States?}   &  \makecell[c]{North Korea} &  \makecell[c]{Temporally \\ambiguous}\\
    \hline
    6  &  \makecell[c]{What was the unemployment rate in Jordan?}  &  \makecell[c]{20 percent}   &  \makecell[c]{Temporally \\ambiguous}\\
    \hline
\end{tabular}
\end{table}

\end{document}